\documentclass[a4paper,11pt]{article}

\usepackage{breakurl}
\usepackage[breaklinks=true]{hyperref}
\usepackage{breakcites}

\usepackage{amssymb,amsmath,array,bm}
\usepackage{booktabs}

\usepackage{tikz}
\usetikzlibrary{arrows}

\usepackage{pgfplots}
\pgfplotsset{compat=1.15}
\usetikzlibrary{arrows}

\usepackage[a4paper, left=3cm, right=2.5cm, top=3cm, bottom=3cm]{geometry}

\usepackage{authblk}
\usepackage{natbib}
\usepackage{doi}

\usepackage[utf8]{inputenc}
\usepackage[T1]{fontenc}
\usepackage[english]{babel}
\usepackage{csquotes}

\usepackage{graphicx}
\usepackage{booktabs}

\title{SCRIPT: Implementing an Intelligent Tutoring System for Programming in a German University Context}

\author[1]{Alina Deriyeva}
\author[1,2]{Jesper Dannath}
\author[1]{Benjamin Paaßen}
\affil[1]{Faculty of Technology, Bielefeld University}

\date{preprint of \cite{SCRIPT2025} as provided by the authors}

\begin{document}

\maketitle

\pagestyle{myheadings}
\markright{preprint of \cite{SCRIPT2025} as provided by the authors}

\footnotetext[2]{Jesper Dannath and Alina Deriyeva both contributed equally.}

\setcounter{footnote}{2}

\begin{abstract}
Practice and extensive exercises are essential in programming education. Intelligent tutoring systems (ITSs) are a viable option to provide individualized hints and advice to programming students even when human tutors are not available. However, prior ITS for programming rarely support the Python programming language, mostly focus on introductory programming, and rarely take recent developments in generative models into account. We aim to establish a novel ITS for Python programming that is highly adaptable, serves both as a teaching and research platform, provides interfaces to plug in hint mechanisms (e.g.\ via large language models), and works inside the particularly challenging regulatory environment of Germany, that is, conforming to the European data protection regulation, the European AI act, and ethical framework of the German Research Foundation. In this paper, we present the description of the current state of the ITS along with future development directions, as well as discuss the challenges and opportunities for improving the system.

\end{abstract}

\section{Description of the AIED implementation, practice or policy}

Intelligent Tutoring Systems (ITSs) have shown their effectiveness in educational scenarios \cite{kulik2015,vanlehn_2011}. They can help to bridge the gap between demand and supply in education, particularly for skills which require extensive exercise, as well as hints and feedback for these exercises, such as programming \cite{mcbroom2021,Crow2018}. We consider the educational context of advanced undergraduate courses, where we introduce Python as a new programming language to experienced computer science students. While we focus on data science education, the system should be sufficiently flexible to support a wide variety of Python programming courses; and teachers should be able to simply upload new course material. In this educational context, we aim to set up an ITS that serves not only as an educational tool, but also as a research platform to develop and investigate different components for ITSs, such as knowledge tracing models or hint and feedback policies based on large language models (LLMs) \cite{roest_2024}. Finally, we aim for a system that is compatible with the current, complex regulatory environment of a German university, including the European data protection regulation (GDPR) \cite{bai2024re,Portela2024}, the European AI act \cite{Mueck2025}, and the ethical framework of the German Research Foundation \cite{DFG2022}. Starting from these requirements, we found that no pre-existing ITS fits our needs sufficiently.

Accordingly, we are developing SCRIPT (Step-based Coding for Research and Intelligent Programming Tutoring), a novel ITS to support advanced undergrad computer science students in their acquisition of data science and machine learning skills\footnote{The source code is available at \url{https://gitlab.ub.uni-bielefeld.de/publications-ag-kml/script/}}. The ITS functions both as a teaching and as a research tool by enabling to record fine-grained, keystroke-level data and is easily adaptable to novel learner models and pedagogical models as well as randomized experiments. Finally, the ITS is designed to respect the regulatory requirements of Bielefeld University (Germany). 
The system was already offered as voluntary exam preparation for an "Introduction to Data Mining" course. As a research tool, the ITS has already supported data collection for learner model and feedback research \cite{Dannath2024,Deriyeva2024,Dannath2025EDM}.

\subsection{System Architecture}

\begin{figure}
\centering
\includegraphics[width=12cm]{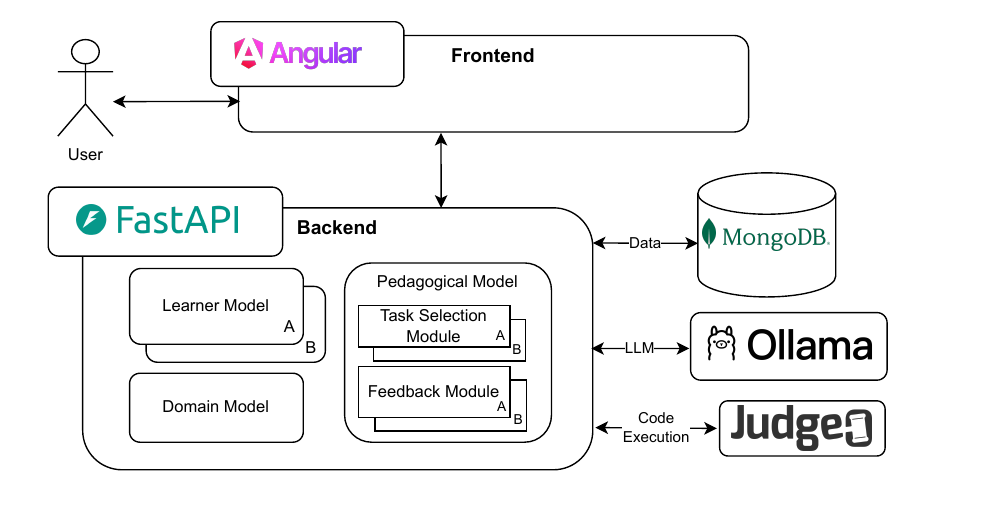}
\caption{Technical architecture of the ITS.}
\label{fig:its_structure}
\end{figure}

We adopt a 4-models approach for this ITS \cite{Sottilare2013} including a Pedagogical Model, a Learner Model, a Domain Model, and a User Interface. Figure~\ref{fig:its_structure} shows how these Models are integrated in the technical architecture of the system. The system is provided as a web-application which is hosted in Docker containers on the university's own server infrastructure. The Angular-based\footnote{\url{https://angular.dev/}} frontend (UI) handles all interactions with the user, while all processing is being done server-side. All other system models are encapsulated in the backend. We choose FastAPI\footnote{\url{https://fastapi.tiangolo.com/}} for the implementation in order to be able to utilize Python's data science and machine learning stack and use MongoDB\footnote{\url{https://www.mongodb.com/}} as a database in order to store Python objects directly as JSON-documents. The system models are designed to be easily replaceable, the learner model and pedagogical model can be selected individually within each course or can be randomly assigned to users for A/B-testing. 
All user code is executed with Judge0 \cite{Dosilovic2020} to ensure security and allow for fixed compute-resource allocation as well as timeouts. Finally, we use Ollama\footnote{https://ollama.com/} as a hosting service for open-weight language models, which we use for generating programming hints.

\begin{figure}
\centering
\includegraphics[width=9cm]{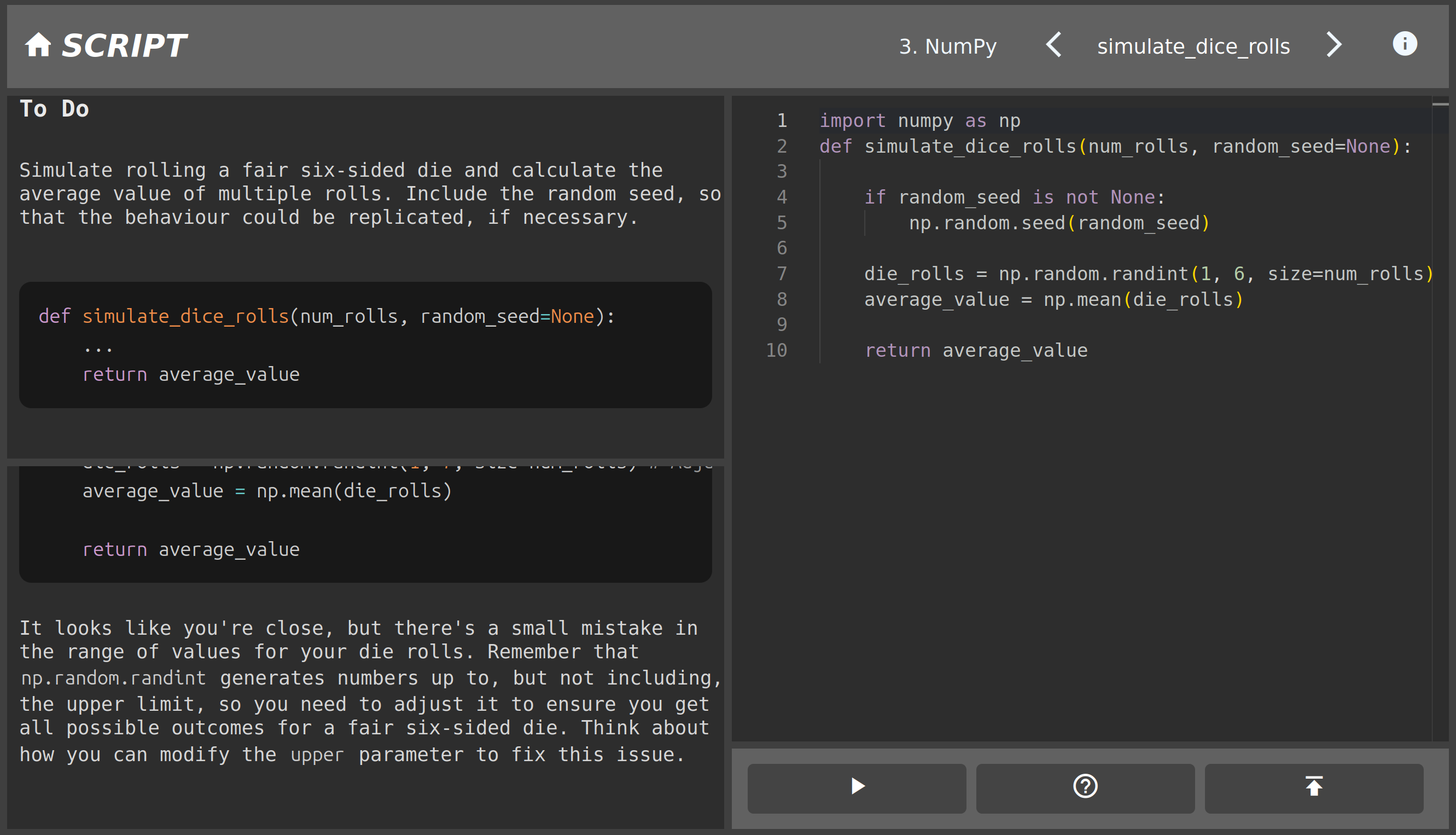}
\caption{Screenshot of the current state of the system's UI.}
\label{fig:its_ui}
\end{figure}

\textbf{User Interface:} The User Interface (UI) connects the other modules to the learner by enabling users to interact with the learning contact and displaying tasks as well as hints to the learner. 
Figure~\ref{fig:its_ui} presents the current interface of the system. The left upper part contains task descriptions, the left bottom part is reserved for displaying different types of feedback (either upon a submission, or whenever a hint is requested).

\textbf{Domain Model:} The domain model contains all tasks as well as information that connects them, like a competency model or a list of knowledge components per task (Q-matrix) \cite{Barnes2005}. Also, it manages and distributes all information related to the difficulty of tasks or knowledge components.

\textbf{Learner Model:} The Learner Model includes all information on the individual learner’s competency and mastery of knowledge components. This information is derived via the learner’s UI actions and a knowledge tracing algorithm \cite{abdelrahman_knowledge_2022}.

\textbf{Pedagogical Model:} Finally, we adopted the differentiation of inner and outer loop \cite{VanLehn2006} for the core ITS features. The outer loop is responsible for task selection and the inner loop is responsible for generating feedback to steps inside programming tasks. 

\begin{figure}
\centering
\includegraphics[width=12cm]{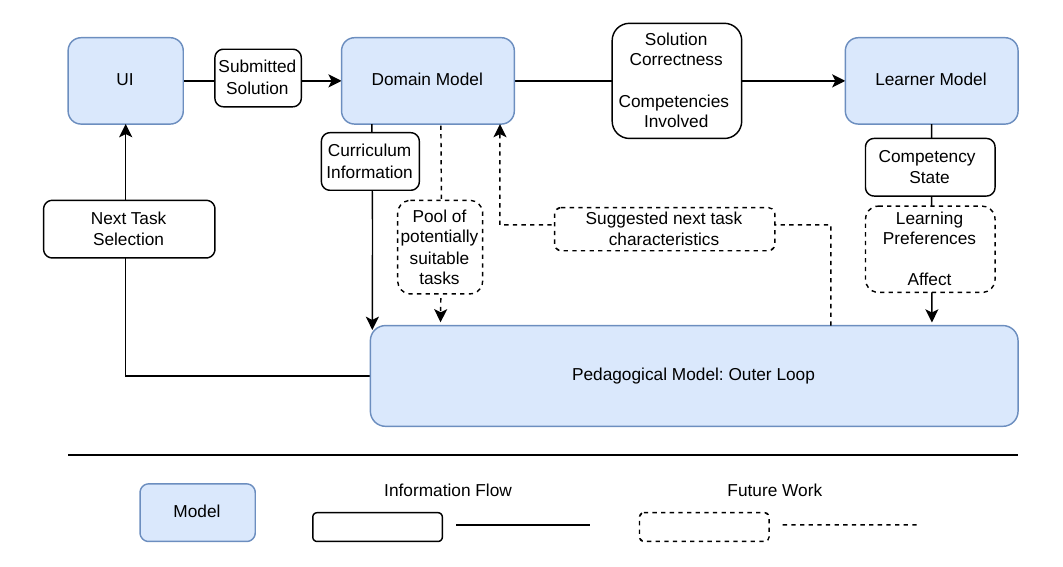}
\caption{Schema of the outer loop of the ITS.}
\label{fig:outer_loop}
\end{figure}

More specifically, the outer loop receives information provided by the domain model (esp.\ task difficulty and task-to-skill relations), and the learner model (esp.\ ability in each skill). Figure~\ref{fig:outer_loop} shows the respective flow (currently implemented parts and potential directions for future work are indicated in the figure) of information involved in the outer loop. The baseline version of the outer loop is a pre-defined curriculum.

\begin{figure}
\centering
\includegraphics[width=9cm]{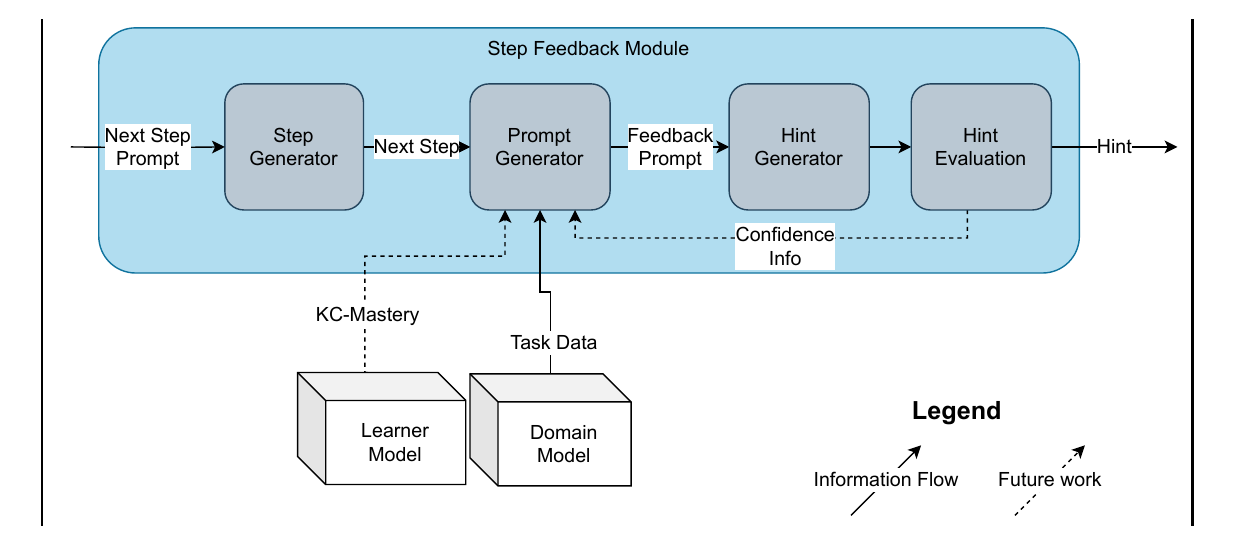}
\caption{Schema of the inner loop of the ITS.}
\label{fig:inner_loop}
\end{figure}

For the inner loop (Figure~\ref{fig:inner_loop}), we use next-step hints as a framework for step-based feedback \cite{mcbroom2021}. This allows to focus on different aspects of feedback, like feedback on mistakes, potential optimizations of user code, or missing pieces of knowledge which can all be the basis for a suggested next step. We decompose the process of hint generation into two modular sub-components, the "Step Generator" and the "Hint Generator", similar to prior work in the ITS literature \cite{birillo_2024,VanLehn2006}. The "Step Generator" predicts the next program state of the student in the direction of the correct solution. The "Hint Generator" has the goal to translate the predicted step into a textual hint that closes the gap between the learner's current knowledge and the required knowledge to continue with the task, without disclosing the complete step. Therefore, the feedback generator should utilize additional information about learner competency as well as task requirements. In a first iteration, we implemented the Step Generator and Hint Generator using different prompt templates for the in-context prompting of LLMs (Figure~\ref{fig:inner_loop}; all prompts are available in the source code repository), which already incorporate contextual information on tasks from the domain model. 

As the system is still in its alpha stage, only contains part of the features described in Figures \ref{fig:outer_loop} and \ref{fig:inner_loop} are already implemented. In the future, contextual information from the learner model will be incorporated, as well. Additionally, we plan to include revision steps based on an evaluation of the generated feedback via a certainty metric. In further iterations, the certainty information of the generated LLM output will be used to further improve the feedback generation.

\subsection{Context of the ITS integration}

\textbf{Educational context:}
SCRIPT is intended to support learning Python by computer science students at the Faculty of Technology at Bielefeld University, Germany. In particular, it was used in context of introductory Python sessions for the courses "Introduction to Machine Learning" and "Introduction to Data Mining", both advanced undergraduate courses for computer science students. Additionally, it was offered as a supplementary exam preparation tool for the "Introduction to Data Mining" course, where the topics were more advanced (refer to Table~\ref{table:exam_prep_tasks} for a full list of tasks). 
The current target group predominantly consists of advanced undergraduate students (third year) and early postgraduate students (first year).

\begin{table}
\centering
\renewcommand{\arraystretch}{1.0}  
\begin{tabular}{ll}
Subdomain & Task Names \\
\cmidrule(r){1-1} \cmidrule{2-2}
Probability Theory & [bayes, dice, independence, mean\_and\_std] \\
Statistical Testing & [pearson, pooled\_variance, wilcoxon] \\
PCA & [principal\_components, factor\_sampling, variance\_coverage] \\ 
Clustering & [kmeans, ward\_linkage, optimal\_gmm\_clusters] \\ 
Logistic Models & [icc, pfa\_inference, IRT\_logreg] \\ 
Markov Models & [most\_likely\_to\_follow, hmm\_sampling, bkt] \\ 
Deep Models & [dkt\_input, training\_loop, vae\_sampling] \\ 
Recommenders & [ndcg\_at\_k, user\_similarity\_explicit, user\_similarity\_cosine]
\end{tabular}
\caption{Tasks with the exam preparation curriculum for Introduction to Data Mining course.}
\label{table:exam_prep_tasks}
\end{table}

\textbf{Regulatory context:}
Within the context of a German university, heightened regulation has to be taken into account. First, because the system processes data of students that could be tied back to individuals (such as IP addresses), it falls within the scope of the EU general data protection regulation (GDPR) \cite{bai2024re,Portela2024}. Therefore, we strictly separated IP address logs from the actual ITS and instructed students to use pseudonymous usernames in order to avoid any coupling between their data stored in the ITS and their identity. Further, we ensured that no user e-mail addresses are stored beyond the registration process. Our strict adherence to data privacy also meant that we were not yet able to implement the system for mandatory task submissions because that would have required to couple submissions to individual student identities. While this is not strictly forbidden under GDPR, it would mean heightened privacy risks which we wanted to avoid. As such, using the tool had to remain entirely voluntary.

Second, because we want to use the ITS not only as a teaching but also as a research tool, we set the system up to record fine-grained (keystroke-level) data within the ITS, to allow detailed evaluation of the submitted code \cite{jeuring_2022}.
However, ethical reasons as well as GDPR forbid us from recording data without explicit consent \cite{DFG2022}. As such, we enabled users to use the ITS as a pure teaching tool but also to provide explicit, additional consent to have their data recorded and stored for research purposes. We highlight that fine-grained consent acquisition is a research topic in itself that is far from resolved \cite{Judel2024}. Further, because one of the courses is taught by the researchers themselves, it had to be ensured that the researchers had no chance to identify which students used the system -- otherwise, students may feel pressured to participate in the research purely to ingratiate themselves with the teachers. Accordingly, in addition to all aforementioned privacy measures, we ensured that the teachers had no access to the servers during the semester but only after another manual anonymization step of all data had been completed and all grades had already been finalized.

Finally, some applications of AI in education are regarded as high risk in the upcoming European AI act \cite{Mueck2025}. While the precise implications of the EU AI Act for AI in education are not yet fully clear, we aimed to be prepared for the novel regulatory environment and avoided invoking any commercial LLM interface. Instead, we opted for a LLama-70b model, hosted by us or a direct research partner. Regulatory constraints aside, we argue that such a setup is more suited to reduce the environmental impact of ITS operation (due to a smaller model size compared to very large LLMs), enhances scientific reproducibility (because we can exactly specify the model version we are using), and decreases dependency upon commercial LLM providers. Most importantly, self-hosted LLMs can reduce risks in terms of privacy and data security.

\section{Reflection of the challenges and opportunities associated with the implementation}        

\textbf{Opportunities:}

The system is in an ongoing improvement cycle, where we gradually add features to improve the quality of the system. That is, each cycle would consist of integration of new features (in the system or curriculum-wise) and a following semester-long testing in a real world setting. At present, the ITS is only used in a narrow, short-term setting (Python introduction and exam preparation). In the future, our intention is to roll the ITS out as a course-long support tool for studying and solving homework assignments.
Further, we also regard the ITS as a training and development tool in study projects, where students can develop novel features and thereby improve the system, themselves.

The ITS also serves as a research platform, in the sense that it allows testing different educational strategies and tools, and allows collecting detailed data about learning interactions of the users (currently at the key-stroke level). This includes, but is not limited to: learner models (what models of knowledge tracing \cite{abdelrahman_knowledge_2022} represent the actual performance of users the best), outer loop (what task/intervention suggestion models best support skills acquisition and promote the efficiency of learning), feedback generation (What hints to give in which situation/to different users and how hints should be presented to users in order to achieve ideal learning gains).

\textbf{Challenges:}
%
Developing an ITS from scratch proves to be challenging. As development is mainly guided by two PhD students, the sheer workload of development tasks provides a strain on the time that can be devoted to research. Similarly, keeping up with constantly progressing technologies requires a considerable amount of time. Moreover, as the system is work-in-progress, IT security vulnerabilities are likely: for example, there was a significant vulnerability in the system before the incorporation of judge0 for code execution, which was pointed out by a couple of enthusiastic students. 

Complying with the research ethics norms and installing the procedures mentioned in the previous section proved to be additional, substantial challenges. In fact, the roll-out of a semester-long teaching plan via the ITS had to be postponed for a year due to a need to find suitable procedures that comply with all regulations and ensure their approval by the respective committees. While we had anticipated some of these challenges, the sheer complexity of the compliance efforts surprised us. We are convinced that, in the end, these efforts changed our system for the better. However, anyone administering AI systems for education in an EU (in particular, German) context should anticipate this additional effort.
   
\section{Description of future steps}  

The main next milestone is related to the incorporation of the ITS in a semester-long curriculum to support the "Introduction to Data Mining" course. This milestone aims to accomplish both educational (longitudinal support of knowledge acquisition) and research (investigating the systems effectiveness in pre- and post-test setups) goals. However, there are several additional features that we plan to incorporate into the ITS.

\textbf{Authoring interface:} The current version of the ITS requires manual task file uploads to change the teaching content. Instead, we plan to provide a new component which enables teachers to directly edit tasks, similar to the idea of authoring tools in tutoring systems \cite{dermeval_2018}.

\textbf{Pedagogical model:} We aim to test different recommendation methods for next tasks to investigate their impact onto learning efficiency. Another goal is to enrich the ITS with intervention techniques when a learner struggles to progress within their curriculum; for example, recommending additional materials/hints for a specific skill/topic. This recommendations would follow a pathway similar to the one shown in the Figure~\ref{fig:outer_loop}. All these variations of the outer loop will also have to be validated in empirical studies. 
Within the inner loop, the ITS currently allows to request an LLM-generated feedback to the current state of the code of the learner. As displayed in Figure~\ref{fig:inner_loop}, this feature is planned to be further refined in order to increase feedback reliability. Furthermore, we are investigating the ideal step granularity for next step hints and will incorporate our research results into the ITS in the future \cite{Dannath2025EDM}. 

\textbf{Learner Model:} To further support the users of the ITS, we intend to implements a dashboard that would present the findings of the learner model to the users. Additionally, we want to improve the explainability of the learner model's results and employ methods to facilitate interpretability of the presented information. 

\textbf{Evaluation:} Once the ITS is used as a semester-long support tool, we plan to use A/B testing with pre- and post-tests to compare skill gains of different pedagogical models.

\textbf{Open-Source:} While we released the current source code of the ITS at \url{https://gitlab.ub.uni-bielefeld.de/publications-ag-kml/script/}, the system is still in the alpha stage. As such, we can not yet recommend the system for deployment but only for research use.

\section{Acknowledgements}

We gratefully acknowledge the support of the project “SAIL: SustAInable Lifecycle of Intelligent Socio-Technical Systems” project. SAIL is funded by the Ministry of Culture and Science of the State of North Rhine-Westphalia (Germany) (grant no. NW21-059A). We would also like to express our gratitude to Arno Gaußelmann for his contribution to the development of SCRIPT.

\bibliographystyle{plainnat}
\bibliography{literature}
\end{document}